\documentclass[letterpaper, 10 pt, conference]{ieeeconf}  

\IEEEoverridecommandlockouts                              

\usepackage{xcolor}

\usepackage{graphicx}

\usepackage{gensymb}
\usepackage[hyphens]{url}
\usepackage{multirow}
\usepackage{multicol}
\usepackage{tabularx}
\usepackage{dsfont}
\usepackage{url}
\usepackage{color}
\usepackage{subcaption}
\usepackage{caption}
\usepackage{mathrsfs}
\usepackage{float}
\usepackage{booktabs}
\usepackage{dcolumn}
\usepackage{dingbat}
\let\checkmark\undefined
\usepackage{amssymb}
\usepackage{hyperref}
\usepackage{amsmath}
\usepackage[ruled,linesnumbered]{algorithm2e}

\usepackage{xcolor}
\usepackage[justification=centering]{caption}

\hypersetup{
    colorlinks=true,
    linkcolor=black,
    filecolor=blue,      
    urlcolor=blue,
    pdftitle={Overleaf Example},
    pdfpagemode=FullScreen,
    }

\title{\LARGE \bf
Robust Collaborative 3D Object Detection in Presence of Pose Errors}

\author{Yifan Lu$^{1}$, Quanhao Li$^{2}$, Baoan Liu$^{3}$, Mehrdad Dianati$^{4}$, Chen Feng$^{5}$, Siheng Chen$^{*1,6}$, Yanfeng Wang$^{1,6}$ 
    \thanks{$^{1}$ Cooperative Medianet Innovation Center (CMIC), Shanghai Jiao Tong University, China.
            {\tt\small \{yifan\_lu, sihengc, wangyanfeng\}@sjtu.edu.cn}.}
    \thanks{$^{2}$ Nanjing University, China.
            {\tt\small quanhaoli2002@163.com}}
    \thanks{$^{3}$ Meta Reality Labs, USA.
            {\tt\small baoanliu@meta.com}}
    \thanks{$^{4}$ University of Warwick, UK.
            {\tt\small M.Dianati@warwick.ac.uk}}
    \thanks{$^{5}$ New York University, USA.
            {\tt\small cfeng@nyu.edu}}
    \thanks{$^{6}$ Shanghai AI laboratory, China.}
    \thanks{$^{*}$ Corresponding author.}
}

\begin{document}

\maketitle
\thispagestyle{empty}
\pagestyle{empty}

\begin{abstract}

Collaborative 3D object detection exploits information exchange among multiple agents to enhance accuracy of object detection in presence of sensor impairments such as occlusion. However, in practice, pose estimation errors due to imperfect localization would cause spatial message misalignment and significantly reduce the performance of collaboration. To alleviate adverse impacts of pose errors, we propose $\mathtt{CoAlign}$, a novel hybrid collaboration framework that is robust to unknown pose errors. The proposed solution relies on a novel agent-object pose graph modeling to enhance pose consistency among collaborating agents. Furthermore, we adopt a multi-scale data fusion strategy to aggregate intermediate features at multiple spatial resolutions. Comparing with previous works, which require ground-truth pose for training supervision, our proposed $\mathtt{CoAlign}$ is more practical since it doesn't require any ground-truth pose supervision in the training and makes no specific assumptions on pose errors. Extensive evaluation of the proposed method is carried out on multiple datasets, certifying that $\mathtt{CoAlign}$ significantly reduce relative localization error and achieving the state of art detection performance when pose errors exist. Code are made available for the use of the research community at \href{https://github.com/yifanlu0227/CoAlign}{https://github.com/yifanlu0227/CoAlign}.
\end{abstract}

\vspace{-3mm}
\section{INTRODUCTION}
3D object detection has received significant attention in recent years because of its applications in autonomous driving, robotics, metaverse, etc~\cite{lang2019pointpillars,chen20203d,10044977} . Despite the rapid development with large-scale datasets and powerful models, 3D object detection by a single agent suffers from inherent limitations, such as occlusion and distant
objects. By leveraging agent-to-agent communication, such as vehicle-to-everything (V2X) in driving scenarios, multiple agents can share complementary perceptual information with each other, promoting more holistic receptive fields. To achieve such collaborative 3D detection, recent works have contributed high-quality datasets~\cite{Li_2021_RAL, xu2022opv2v, yu2022dairv2x, arnold2021data} and effective collaboration methods~\cite{chen2019f,DBLP:conf/eccv/WangMLYZU20,Li_2021_NeurIPS, arnold2020cooperative}. But there are still numerous challenges in this emerging field, such as communication bandwidth constraints~\cite{DBLP:conf/eccv/WangMLYZU20,Where2comm:22}, latency~\cite{lei2022latency} and adversarial attack~\cite{tu2021adversarial}.  This work focuses on mitigating the negative effect of pose error.


To share valid information with each other, multiple agents need precise poses to synchronize their individual data in a consistent spatial coordinate system, which is a foundation of the collaboration. However, the 6 DoF pose estimated by each agent's localization module is not perfect in practice, causing wanted relative pose errors. Such relative pose errors would fundamentally reduce the collaboration quality~\cite{DBLP:conf/eccv/WangMLYZU20}. To address this issue, previous works consider various methods to promote pose robustness~\cite{glaser2021overcoming, xu2022v2xvit,pmlr-v155-vadivelu21a,9682601}. For example,~\cite{pmlr-v155-vadivelu21a}  designs pose regression module to learn pose correction; and~\cite{9682601} uses 3D point's semantic label to find point-wise correspondence. However, these methods require the ground-truth poses in training data. Although pose errors in training data could be corrected offline, this labeling process could be expensive and imperfect.

\begin{figure}[t]
    \centering{\includegraphics[width=0.99\linewidth]{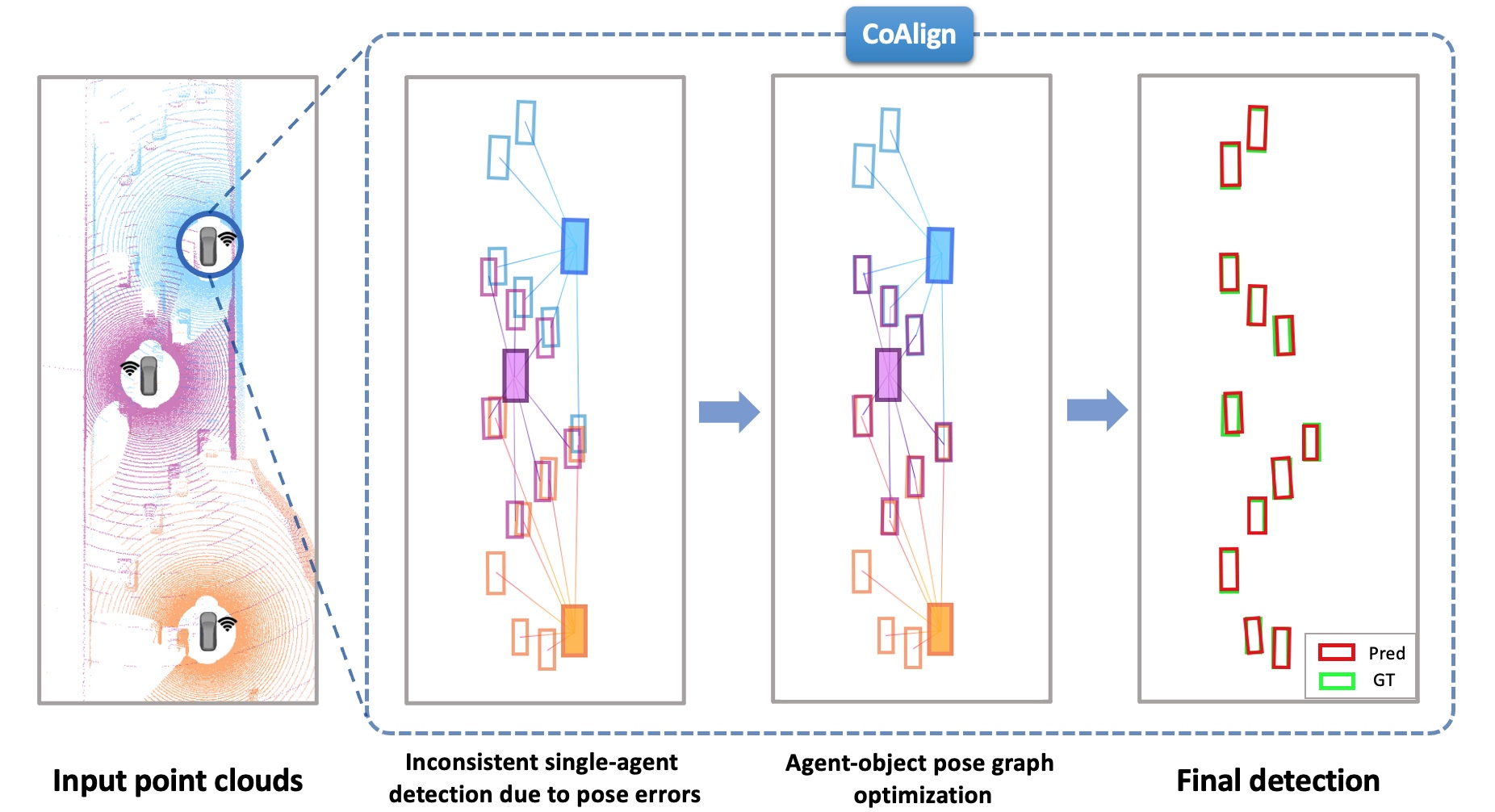}}
    \caption{Based on agent-object pose graph optimization, our $\mathtt{CoAlign}$ reduces relative pose errors and improves detection robustness.}
    \vspace{-7mm}
    \label{fig:teaser}
\end{figure}

Motivated by this limitation, we propose a novel hybrid collaboration framework $\mathtt{CoAlign}$, which enables multiple agents to share both intermediate features and single-agent detection results. $\mathtt{CoAlign}$ can handle arbitrary pose errors \textbf{without any accurate pose supervision} in the training phase. To realize this, the core idea is to leverage novel agent-object pose graph optimization to align the relative pose relations between agents and detected objects in the scene, promoting pose consistency. Here, pose consistency means that the poses of an object should be consistent from multiple agents' perspectives. Since our agent-object pose graph does not use any training parameters in the optimization process, this method has strong generalization capability to adapt to arbitrary levels of pose errors. To effectively alleviate the impact of pose error, we further consider a multiscale intermediate fusion strategy, which comprehensively aggregates collaboration information at multiple spatial scales. 

We conduct extensive experiments for the task of LiDAR-based 3D object detection on both simulation and real-world datasets, including OPV2V~\cite{xu2022opv2v}, V2X-Sim 2.0~\cite{Li_2021_RAL} and DAIR-V2X~\cite{yu2022dairv2x}. The results show that i) the proposed agent-object pose graph optimization achieves 75\% relative pose error reduction (measured at the median value); and ii) the proposed $\mathtt{CoAlign}$ achieves at least 12\% performance improvement in the task of collaborative 3D object detection with pose error existence, compared with other methods.

In summary, the main contribution of this work are:

$\bullet$ We propose $\mathtt{CoAlign}$, a novel robust multi-agent collaborative LiDAR-based 3D detection framework that deals with unknown pose errors in both training and testing phases;

$\bullet$ We propose novel agent-object pose graph modeling and optimization, which corrects relative poses among multiple agents by promoting the consistency of relative poses;

$\bullet$ We conduct extensive experiments to validate that $\mathtt{CoAlign}$ can greatly reduce the relative pose error and achieve more accurate and robust 3D detection performances.


\section{Related Works}

\subsection{Collaborative Perception}
As a recent application of multi-agent systems to perception tasks, collaborative perception is emerging{~\cite{xu2022cobevt,xu2022v2xvit,chen2019f,DBLP:conf/eccv/WangMLYZU20}}. To support this area of research, there is a surge of high-quality datasets, including V2X-Sim~\cite{Li_2021_RAL}, OPV2V~\cite{xu2022opv2v}, Comap~\cite{comap} and DAIR-V2X~\cite{yu2022dairv2x}. Based on those datasets, numerous methods have been proposed to handle various practical issues. For example, to achieve a better tradeoff between perception performance. DiscoNet~\cite{Li_2021_NeurIPS} proposes a teacher-student knowledge distillation framework in training stage to harness the more comprehensive perceptive information from early fusion. To handle communication latency, SyncNet~\cite{lei2022latency} leverages feature-attention symbiotic estimation and time modulation techniques. For system security, ~\cite{tu2021adversarial} investigate adversarial attacks in the setting of multi-agent communication. In this work, we specifically consider the robustness towards localization error.



\subsection{Localization Issue in Collaborative Perception}
Precise localization of each agent is a foundation of multi-agent collaboration{~\cite{pmlr-v155-vadivelu21a, glaser2021overcoming, xu2022v2xvit, 9682601,yuan2022leveraging}}. Inaccurate poses would cause misalignment and inconsistency in collaboration, resulting in worse perception performance than single-agent perception. To gain resistance to localization noises, previous works consider two main approaches: supervised training or robust network design. The first approach introduces additional supervision to empower the network being aware of the pose errors. For example, V2VNet (robust)~\cite{pmlr-v155-vadivelu21a} designs pose regression, global consistency and attention aggregation module to correct relative poses and concentrate on neighbor with less pose error; MASH~\cite{glaser2021overcoming} builds a similarity volume and explicitly learns the pixel to pixel correspondence to avoid using noisy pose in inference. The second approach focuses on designing robust frameworks or network architectures. For example, V2X-ViT~\cite{xu2022v2xvit} uses multi-scale window attention to capture features in various ranges; and FPV-RCNN~\cite{9682601} infers the semantic label of keypoints and finds correspondences between agents to correct relative poses. However, most of these previous methods require ground-truth poses for supervision in the training stage, which makes them less practical. In comparison, the proposed $\mathtt{CoAlign}$ does not make any specific assumption on the pose noises and does not require any ground-truth poses in training and testing phases with the help of agent-object pose graph optimization.
 
\subsection{Simultaneous Localization and Mapping (SLAM)}
In robotics, SLAM\cite{taketomi2017visual,aulinas2008slam,saputra2018visual,zou2019collaborative} aims to estimate the map of the environment while localizing the agent in the map, which is being built. A standard SLAM system consists of a frontend and a backend. The frontend takes in raw sensor data and transforms it into an intermediate representation, such as constraints of an optimization problem; and the backend solving it to estimate the positional state of agent and landmarks. As a general formulation of SLAM, graph-based SLAM~\cite{lu1997globally,grisetti2010tutorial} builds a pose graph whose nodes are the poses of the agent at various time stamps and edges represents spatial constraints between the poses. Solving graph-Based SLAM is to find a node configuration that minimize the error introduced by the constraints.  The proposed agent-object pose graph is similar to the pose graph in graph-based SLAM; however, the former considers the pose of an object from the perspectives of multiple agents at the same time stamp, while the latter one considers the pose of the same agent across multiple time stamps.





\begin{figure*}[h]
\centering
\centering{\includegraphics[width=0.99\linewidth]{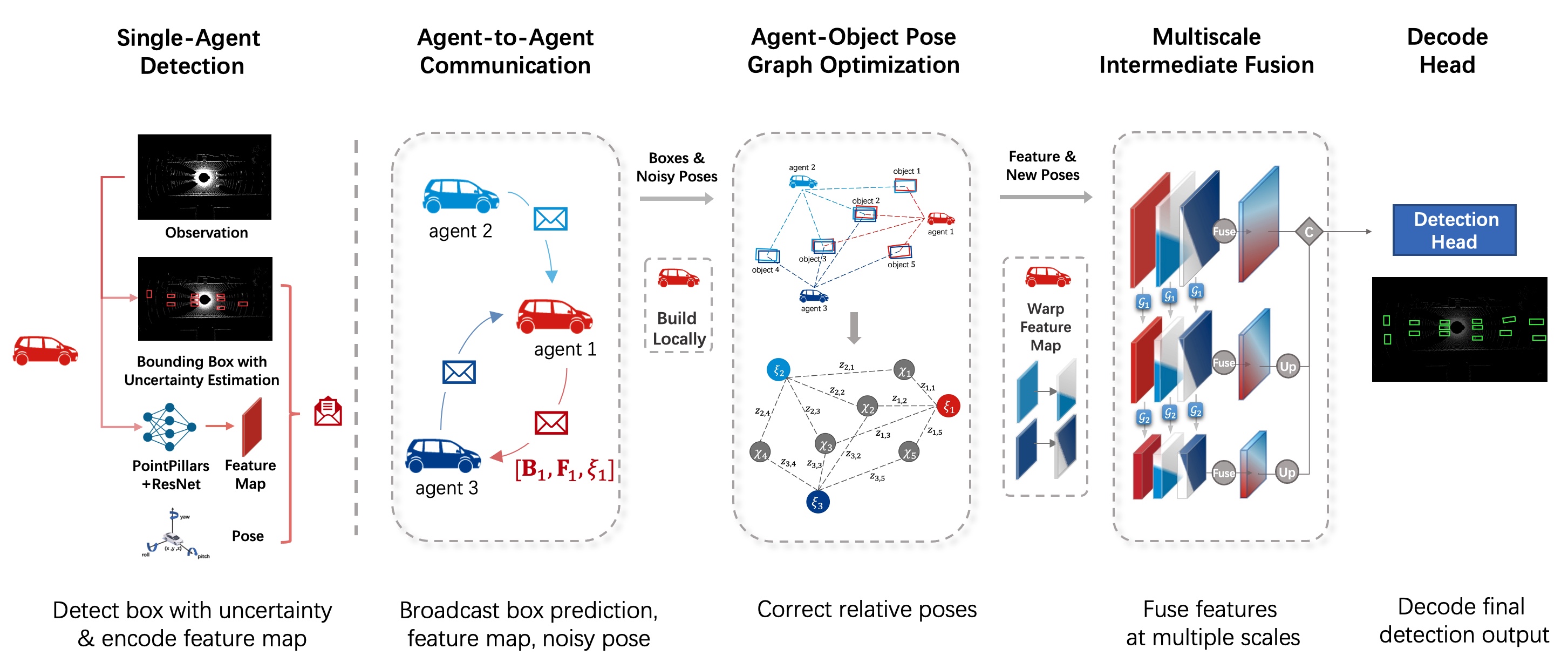}}
\caption{Overview of $\mathtt{CoAlign}$.
Before communication, each agent uses input raw observation to i) predict the bounding box with uncertainty estimation ii) generate intermediate features. Agents pack i) and ii) with measured poses and broadcast to other agents. By aggregating incoming messages, each agent builds the agent-object pose graph and locally optimizes relative poses. Corrected poses are utilized for warping feature maps to the ego coordinate, which goes through a multiscale intermediate fusion module afterward. Finally, fused features are decoded into the final detection output.}
\label{fig:overview}
\vspace{-3mm}
\end{figure*}

\section{Problem Formulation}
Consider $N$ agents in the scene. Each agent has the ability to perceive, communicate and detect. The goal is to reach a better 3D detection ability of each agent by distributedly collaborating with each other. There are three types of collaboration in the previous literature: early collaboration\cite{arnold2020cooperative, xu2022opv2v}, which transmits raw observation data, intermediate collaboration\cite{Li_2021_RAL,xu2022v2xvit}, which transmits intermediate features, and late collaboration\cite{arnold2020cooperative, xu2022opv2v}, which transmits detection output. Many previous works focus on intermediate collaboration as transmitting features is more compact and effective. Let $\mathbf{O}_{i}$ and $\mathbf{B}_i$ be the perceptual observation and the  detection output of the $i$th agent, respectively. For the $i$th agent, a standard 3D object detection based on  intermediate collaboration  works as follows:
\begin{subequations}
    \begin{eqnarray}
    \label{eq:encoder}
    \vspace{-5mm}
    &&\mathbf{F}_i = f_{\rm encoder} \left(\mathbf{O}_{i} \right),
    \\
    \label{eq:transform}
    &&\mathbf{M}_{j \rightarrow i} = f_{\rm transform} \left(
    \xi_i , (\mathbf{F}_j, \xi_j)\right),
    \\
    \label{eq:fusion}
    &&\mathbf{F}'_i = f_{\rm fusion} \left( \mathbf{F}_i, \{ \mathbf{M}_{j \rightarrow i} \}_{j=1,2, \cdots,N} \right),
    \\
    \label{eq:decoder}
    &&\mathbf{B}_i = f_{\rm decoder} \left( \mathbf{F}'_i \right),
    \end{eqnarray}
\end{subequations}
where $\mathbf{F}_i$ is the feature extracted from the $i$th agent's observation, $\xi_i = (x_i,y_i,z_i,\theta_i, \phi_i, \psi_i)$ is the 6DoF pose of the $i$th agent with $\theta_i, \phi_i, \psi_i$ the  yaw, pitch and roll angle, $\mathbf{M}_{j \rightarrow i}$ is the message transmitted from the $j$th agent to the $i$th agent, whose coordinate space is aligned with $\mathbf{F}_i$ through pose transformation, and $\mathbf{F}'_i$ is the aggregated feature of the $i$th agent after fusing other agents' messages. Note that: i) before pose transformation, the direct message transmitted from the $j$th agent to the $i$th agent is $(\mathbf{F}_j, \xi_j)$, including both the feature and the pose; and ii) single-agent detection only considers Steps~\eqref{eq:encoder} and~\eqref{eq:decoder}  with $\mathbf{F}'_i = \mathbf{F}_i$.

In practice, each 6DoF pose $\xi_i$ is estimated by the localization module is usually noisy. Then, after pose transformation in Step~\eqref{eq:transform}, each message $\mathbf{M}_{j \rightarrow i}$ would have different intrinsic coordinate systems, resulting in fusion misalignment in Step~\eqref{eq:fusion} and unsatisfying detection output in Step~\eqref{eq:decoder}. The goal of this work is to minimize the effect of pose errors by introducing additional pose correction before Step~\eqref{eq:transform}.






\section{Pose-Robust Collaborative 3D Detection}

\subsection{Overall Architecture}
Now we propose $\mathtt{CoAlign}$, a novel hybrid collaboration framework, which combines intermediate and late collaboration strategy. Compared with intermediate fusion, this hybrid collaboration can leverage those bounding boxes detected by agents to be the scene landmarks and correct the relative poses between agents. Mathematically, for the $i$th agent, $\mathtt{CoAlign}$ works as:
\begin{subequations}
    \begin{eqnarray}
    \label{eq:coalign_encoder}
    \mathbf{F}_i, \mathbf{B}_i & = & f_{\rm detection} \left(\mathbf{O}_{i} \right),
    \\
    \label{eq:coalign_correction}
    \{ \xi'_{j \rightarrow i} \}_j & = & f_{\rm correction} \left(
    \{\mathbf{B}_j, \xi_j \}_{j=1,2,\cdots, N} \right),
    \\
    \label{eq:coalign_transform}
    \mathbf{M}_{j \rightarrow i} & = & f_{\rm transform} \left( \mathbf{F}_j, \xi'_{j \rightarrow i}\right),
    \\
    \label{eq:coalign_fusion}
    \mathbf{F}'_i & = & f_{\rm fusion} \left( \{ \mathbf{M}_{j \rightarrow i} \}_{j=1,2, \cdots,N} \right),
    \\
    \label{eq:coalign_decoder}
    \mathbf{B}'_i  & = & f_{\rm decoder} \left( \mathbf{F}'_i \right),
    \end{eqnarray}
\end{subequations}
where $\mathbf{B}_i$ and $\mathbf{B}'_i$ are the detection outputs before and after collaboration, respectively, and $\xi'_{j \rightarrow i}$ is the corrected relative pose from $i$th agent's perspective to $j$th agent ($\xi'_{i \rightarrow i}$ is the identity). Step~\eqref{eq:coalign_encoder} is similar to Step~\eqref{eq:encoder}, while Step~\eqref{eq:coalign_encoder} completes the detection by a single agent and further outputs the detected boxes, which will be elaborated in Section~\ref{sec:detection}. As the key step, Step~\eqref{eq:coalign_correction} corrects the relative pose based on all the noisy poses and detected boxes obtained from the other agents; will see details in Section~\ref{sec:correction}. Step~\eqref{eq:coalign_transform} synchronizes other agents' features to the ego pose based on the corrected relative pose. To further ensure the robustness towards pose noises,  Step~\eqref{eq:coalign_fusion} uses a robust multiscale fusion to update the feature, where $\mathbf{M}_{i \rightarrow i} = \mathbf{F}_i$. Finally, Step~\eqref{eq:coalign_decoder} uses fused features to obtain final detections; also see the illustration of the overall architecture in Figure~\ref{fig:overview}.

\subsection{Single-Agent Detection with Uncertainty Estimation}
\label{sec:detection}
Here we elaborate Step~\eqref{eq:coalign_encoder}. The core idea is to generate bounding boxes with the estimated uncertainties, which can serve as the key scene landmarks for the subsequent relative pose correction. To implement the single-agent 3D object detector $f_{\rm detection}(\cdot)$, we can leverage the off-the-shelf designs, such as PointPillars~\cite{lang2019pointpillars}, to produce intermediate features $\mathbf{F}_i$ and estimated bounding boxes $\mathbf{B}_i$ for the $i$th agent. Note that for each bounding box, we also estimate its uncertainty. Since we later rely on those boxes to correct the pose errors, messy detection could cause even worse relative poses. The estimated uncertainty of each box can provide beneficial confidence information to rule out 
undesirable detection.  We parameterize each bounding box with uncertainty as $\mathbf{b} = (\widehat{x},  \widehat{y}, \widehat{z},\widehat{l}, \widehat{w}, \widehat{h}, \widehat{\theta}, \sigma_{x}^2, \sigma_{y}^2, \sigma_{\theta}^2 )$, including the estimated 3D center position, the length, the width, the height, the yaw angle, the center position variance and the angle variance.

To model the center position $x$ (or $y$) as a random variable, we consider a Gaussian distribution with mean $\widehat{x}$ and variance $\sigma_{x}^2$. To supervise the training, we minimize the KL divergence between the estimated Gaussian distribution $\mathcal{N}(\widehat{x},\sigma_{x}^2)$ with the ground-truth delta distribution $\mathbb{P}(x) = \delta(x - x_0)$, where $x_0$ is the groundtruth. Based on the standard derivation, the resulting loss about $x$ is
$
L_{x} \ = \ {\left(\widehat{x} - x_0 \right)^{2}}/{(2 \sigma_{x}^{2})}+ {\log \left(\sigma_{x}^{2}\right)}/{2}.
$

To model the yaw angle, due to its periodicity, we consider a von-Mises distribution with mean $\widehat{\theta}$ and concentration $1/\sigma_{\theta}^2$~\cite{zhong2020uncertainty}. We then minimizing the KL divergence between von-Mises distribution $\mathcal{M}(\widehat{\theta}, 1/\sigma_{\theta}^2)$ and the ground-truth delta function $\mathbb{P}(\theta) = \delta(\theta - \theta_0)$, where $\theta_0$ is the groundtruth, and the corresponding loss about $\theta$ is
$
L_{\theta}=\log I_{0}(\exp (-s))-\exp (-s) \| \cos (\widehat{\theta}-\theta_0)\|,
$
where $s = \log (\sigma_{\theta}^2)$ and $I_0(\cdot)$ is the 0-order modified Bessel function.  To help gradient flow when $s$ is large, we can add another regularization term  $\lambda_{elu} E L U\left(s- c\right)$ with $E L U(\cdot)$ the exponential linear unit (ELU)~\cite{zhong2020uncertainty,clevert2015fast} and $c$ a hyper parameter that controls the effect of ELU.  


Assembling all loss function, the total loss to train single-agent detection jointly with uncertainty estimation is
\begin{equation*}
    L_{total} = L_{cls} + \alpha_{reg} L_{reg} + \alpha_{center} (L_{x} + L_{y}) + \alpha_{\theta} L_{\theta},
\end{equation*}
where $L_{cls}$ is the cross-entropy loss for object classification, $L_{reg}$ is the smooth-L1 loss for box parameter regression, $\alpha_{reg}, \alpha_{center}, \alpha_{\theta}$ are hyperparameters for balancing.


\subsection{Agent-Object Pose Graph Optimization} 
\label{sec:correction}
After the single-agent detection, the $i$th agent shares three types of messages, including i) its pose $\xi_i$ estimated by its own localization module; ii) the bounding boxes detected by the $i$th agent; and iii) its feature map $\mathbf{F}_i$. Note that i) we only need the first two types of messages to correct poses and the third one is used for feature fusion; and ii) since box detection only has yaw angle for rotation measurement, we simplify each pose $\xi_i =(x_i, y_i, \theta_i)$ in 2D space.

To reliably fuse feature maps from other agents, each agent needs to correct the relative poses. To achieve this, the core idea is to align the corresponding bounding boxes of the same object, which are detected by multiple agents. We thus leverage an agent-object pose graph to represent the relations between agents and objects; and then, apply optimization over this graph to achieve pose alignment. Concretely, the $i$th agent receives the messages from all the other agents and builds its internal agent-object pose graph $G(\mathcal{V}^{\rm (agent)}, \mathcal{V}^{\rm (object)}, \mathcal{E})$, which is a bipartite graph to model the relations between agents and detected objects. The node set $\mathcal{V}^{\rm (agent)}$ consists of all the $N$ agents, the node set $\mathcal{V}^{\rm (object)}$ includes all the unique objects, which are obtained via spatially clustering similar boxes received from all the agents, and the edge set $\mathcal{E}$ reflects the detection relationship between agents and objects: when an agent detects a bounding box on an object, we set an edge connecting them. Notably, this agent-object pose graph is built locally at each agent, but it is the same for all the agents.


\begin{figure}[t]
\centering 
\includegraphics[width=0.48\textwidth]{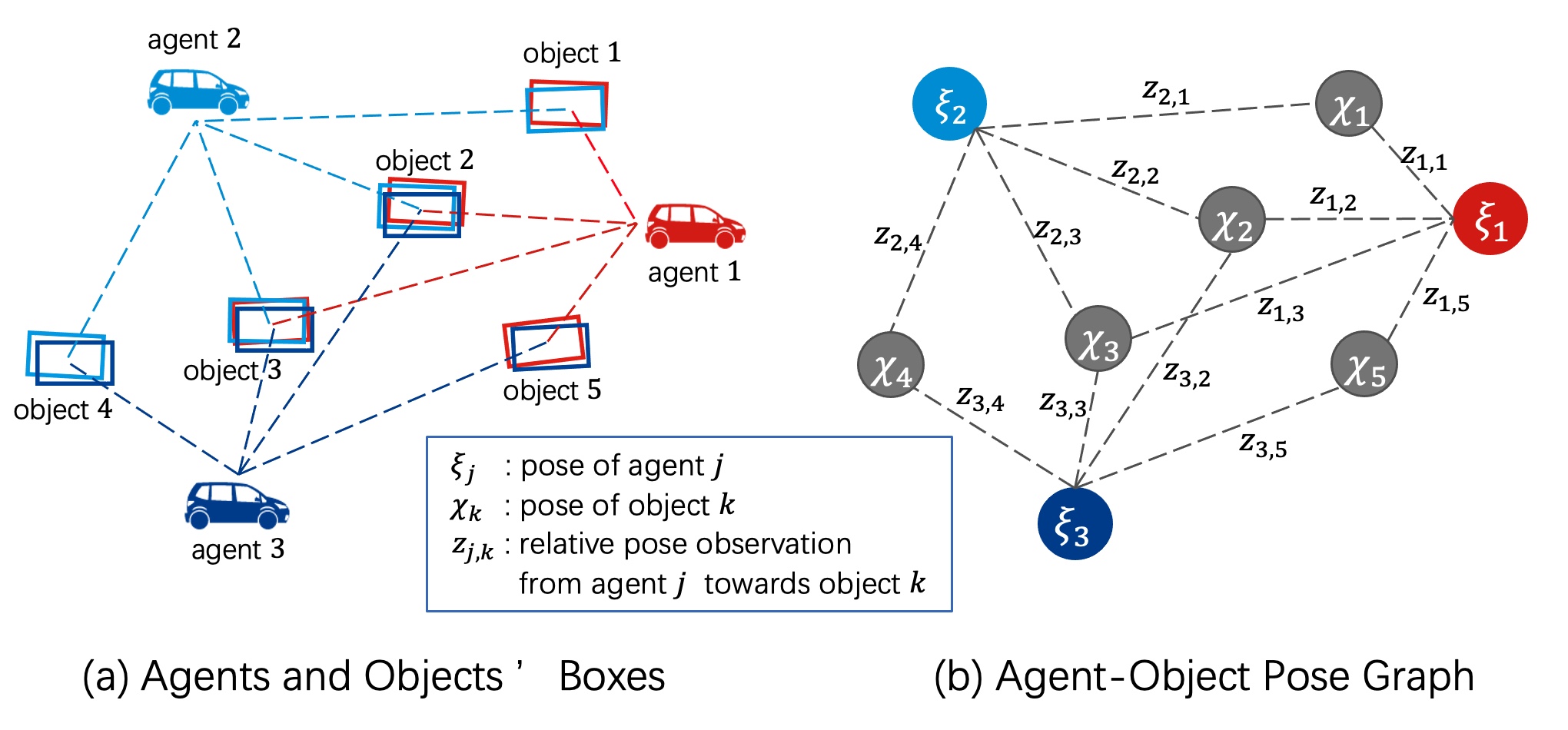} 
\caption{Agent-object pose graph illustration.} 
\vspace{-5mm}
\label{fig:pose_graph} 
\end{figure}

In this agent-object pose graph, each node is associated with a pose. For the $j$th agent in the scene, its pose $\xi_j$ is estimated by its localization module;  for the $k$th object in the scene, its pose $\chi_k$ is sampled from multiple bounding boxes in the same box cluster, which is essentially the unique object detected by multiple agents. Correspondingly, each edge in $\mathcal{E}$ reflects a relative pose between an agent and an object. Let $z_{jk} = (\widehat{x},\widehat{y},\widehat{\theta}) $ be the relative pose of the $k$th object from the perspective of the $j$th agent, which is naturally obtained through the $j$th agent's detection output. Ideally, the pose of an object should be consistent from the views of multiple agents; that is, the pose consistency error vector $\mathbf{e}_{jk} = {z_{jk}^{-1} \circ ( \xi_{j}^{-1} \circ \chi_{k}}) \in \mathbb{R}^3$ should be zero,  where $\circ$ is motion composition operator, equivalent to multiplying their corresponding homogeneous transformation matrices. We denote $\xi^{-1}$ as the inverse pose, equivalent to inverting the corresponding transformation matrix.

To promote this pose consistency, we consider the following optimization problem
\vspace{-1mm}
\begin{equation}
\label{eq:graph_opt}
\{\xi'_{j}, \chi'_{k} \} \ = \  \underset{ \{\xi_j, \chi_{k} \} } \arg \min \mathbf{E}( \mathcal{E} ) \ = \ \sum_{(\xi_j, \chi_{k}) \in \mathcal{E}} \mathbf{e}_{jk}^{T} \mathbf{\Omega}_{jk} \mathbf{e}_{jk},
\vspace{-1mm}
\end{equation}
where $\mathbf{\Omega}_{jk} = {\rm diag}([1/\sigma_{x}^2, 1/\sigma_{y}^2, 1/\sigma_{\theta}^2]) \in \mathbb{R}^{3 \times 3}$ is analogous to information matrix in graph-based SLAM, gauging how we are confident in this measurement. The diagonal elements are from box's uncertainty estimation.

This optimization~\eqref{eq:graph_opt} has a similar formulation with graph-based SLAM, but different meanings: graph-based SLAM aligns the same object across multiple time stamps, while we align the same object at the same time stamps, yet detected by multiple agents. This typical graph optimization~\eqref{eq:graph_opt} can be solved by popular Gaussian-Newton or Levenberg-Marquartdt algorithms~\cite{levenberg1944method}. At iterations, we set the ego agent's pose fixed and update all the other agents' and objects' poses. It usually converges well and fast because a good initial guess $\{\xi_j, \chi_k\}$ has been given. Relative poses are calculated in the end $\xi'_{j \rightarrow i} = \xi_{i}'^{-1} \circ \xi_{j}'$; see more mathematical optimization details in~\cite{grisetti2011g2o}. With the corrected relative pose $\xi'_{j \rightarrow i}$, the $i$th agent can warp the $j$th agent's feature $\mathbf{F}_j$ to its ego pose and obtain $\mathbf{M}_{j \rightarrow i}$, which has the same coordinate system with its ego feature $\mathbf{F}_i$.

\begin{table*}[]
\centering
\resizebox{\textwidth}{!}{%
\begin{tabular}{clcccccccccccc}
\hline
\multicolumn{2}{c}{Dataset} & \multicolumn{4}{c}{\textbf{OPV2V}} & \multicolumn{4}{c}{\textbf{V2X-Sim 2.0}} & \multicolumn{4}{c}{\textbf{DAIR-V2X}} \\ \hline
\multicolumn{2}{c|}{Method/Metric} & \multicolumn{12}{c}{AP@0.5 $\uparrow$} \\ \hline
\multicolumn{2}{c|}{Noise Level $\sigma_t/\sigma_r(m/^{\circ})$} & {\color[HTML]{656565} 0.0/0.0} & 0.2/0.2 & 0.4/0.4 & \multicolumn{1}{c|}{0.6/0.6} & {\color[HTML]{656565} 0.0/0.0} & 0.2/0.2 & 0.4/0.4 & \multicolumn{1}{c|}{0.6/0.6} & {\color[HTML]{656565} 0.0/0.0} & 0.2/0.2 & 0.4/0.4 & 0.6/0.6 \\ \hline
\multicolumn{1}{c|}{} & \multicolumn{1}{l|}{F-Cooper~\cite{chen2019f}} & {\color[HTML]{656565} 0.834} & 0.788 & 0.681 & \multicolumn{1}{c|}{0.604} & {\color[HTML]{656565} 0.679} & 0.634 & 0.568 & \multicolumn{1}{c|}{0.516} & {\color[HTML]{656565} 0.734} & 0.723 & 0.705 & 0.692 \\
\multicolumn{1}{c|}{} & \multicolumn{1}{l|}{V2VNet~\cite{DBLP:conf/eccv/WangMLYZU20}} & {\color[HTML]{656565} 0.935} & 0.922 & 0.884 & \multicolumn{1}{c|}{0.841} & {\color[HTML]{656565} 0.851} & {\color[HTML]{000000} 0.839} & 0.796 & \multicolumn{1}{c|}{0.742} & {\color[HTML]{656565} 0.664} & 0.649 & 0.623 & 0.599 \\
\multicolumn{1}{c|}{} & \multicolumn{1}{l|}{DiscoNet~\cite{Li_2021_NeurIPS}} & {\color[HTML]{656565} 0.916} & 0.906 & 0.884 & \multicolumn{1}{c|}{0.862} & {\color[HTML]{656565} 0.785} & 0.775 & 0.748 & \multicolumn{1}{c|}{0.708} & {\color[HTML]{656565} 0.736} & 0.726 & 0.708 & 0.697 \\
\multicolumn{1}{c|}{\multirow{-4}{*}{\begin{tabular}[c]{@{}c@{}}w/o\\ robust\\ design\end{tabular}}} & \multicolumn{1}{l|}{OPV2V$_{\text{pointpillar}}$~\cite{xu2022opv2v}} & {\color[HTML]{656565} 0.943} & 0.933 & 0.915 & \multicolumn{1}{c|}{0.899} & {\color[HTML]{656565} 0.824} & 0.807 & 0.782 & \multicolumn{1}{c|}{0.757} & {\color[HTML]{656565} 0.733} & 0.723 & 0.708 & 0.697 \\ \hline
\multicolumn{1}{c|}{} & \multicolumn{1}{l|}{MASH~\cite{glaser2021overcoming}} & {\color[HTML]{656565} 0.602} & 0.602 & 0.602 & \multicolumn{1}{c|}{0.602} & {\color[HTML]{656565} 0.643} & 0.643 & 0.643 & \multicolumn{1}{c|}{{\color[HTML]{000000} 0.643}} & {\color[HTML]{656565} 0.400} & 0.400 & 0.400 & 0.400 \\
\multicolumn{1}{c|}{} & \multicolumn{1}{l|}{FPV-RCNN~\cite{9682601}} & {\color[HTML]{656565} 0.858} & 0.817 & 0.591 & \multicolumn{1}{c|}{0.419} & {\color[HTML]{656565} 0.870} & 0.835 & 0.654 & \multicolumn{1}{c|}{0.480} & {\color[HTML]{656565} 0.655} & 0.631 & 0.580 & 0.581 \\
\multicolumn{1}{c|}{} & \multicolumn{1}{l|}{V2VNet$_{\text{robust}}$~\cite{pmlr-v155-vadivelu21a}} & {\color[HTML]{656565} 0.942} & {\color[HTML]{343434} 0.938} & 0.929 & \multicolumn{1}{c|}{{\color[HTML]{000000} 0.918}} & {\color[HTML]{656565} 0.840} & 0.836 & 0.811 & \multicolumn{1}{c|}{0.778} & {\color[HTML]{656565} 0.660} & 0.655 & 0.646 & 0.636 \\
\multicolumn{1}{c|}{} & \multicolumn{1}{l|}{V2X-ViT~\cite{xu2022v2xvit}} & {\color[HTML]{656565} 0.946} & 0.942 & 0.931 & \multicolumn{1}{c|}{0.914} & {\color[HTML]{656565} \textbf{0.881}} & \textbf{0.858} & 0.808 & \multicolumn{1}{c|}{0.759} & {\color[HTML]{656565} 0.704} & 0.700 & 0.689 & 0.678 \\
\multicolumn{1}{c|}{\multirow{-5}{*}{\begin{tabular}[c]{@{}c@{}}w/\\ robust\\ design\end{tabular}}} & \multicolumn{1}{l|}{Ours} & {\color[HTML]{656565} \textbf{0.966}} & {\color[HTML]{000000} \textbf{0.962}} & {\color[HTML]{000000} \textbf{0.958}} & \multicolumn{1}{c|}{{\color[HTML]{000000} \textbf{0.945}}} & {\color[HTML]{656565} 0.858} & 0.852 & {\color[HTML]{000000} \textbf{0.822}} & \multicolumn{1}{c|}{{\color[HTML]{000000} \textbf{0.796}}} & {\color[HTML]{656565} \textbf{0.746}} & {\color[HTML]{000000} \textbf{0.738}} & {\color[HTML]{000000} \textbf{0.720}} & {\color[HTML]{000000} \textbf{0.700}} \\ \hline 
\\ \hline
\multicolumn{2}{c|}{Method/Metric} & \multicolumn{12}{c}{AP@0.7 $\uparrow$} \\ \hline
\multicolumn{2}{c|}{Noise Level $\sigma_t/\sigma_r(m/^{\circ})$} & {\color[HTML]{656565} 0.0/0.0} & 0.2/0.2 & 0.4/0.4 & \multicolumn{1}{c|}{0.6/0.6} & {\color[HTML]{656565} 0.0/0.0} & 0.2/0.2 & 0.4/0.4 & \multicolumn{1}{c|}{0.6/0.6} & {\color[HTML]{656565} 0.0/0.0} & 0.2/0.2 & 0.4/0.4 & 0.6/0.6 \\ \hline
\multicolumn{1}{c|}{} & \multicolumn{1}{l|}{F-Cooper~\cite{chen2019f}} & {\color[HTML]{656565} 0.602} & 0.504 & 0.412 & \multicolumn{1}{c|}{0.376} & {\color[HTML]{656565} 0.489} & 0.434 & 0.379 & \multicolumn{1}{c|}{0.362} & {\color[HTML]{656565} 0.559} & 0.552 & 0.542 & 0.538 \\
\multicolumn{1}{c|}{} & \multicolumn{1}{l|}{V2VNet~\cite{DBLP:conf/eccv/WangMLYZU20}} & {\color[HTML]{656565} 0.740} & 0.686 & 0.586 & \multicolumn{1}{c|}{0.504} & {\color[HTML]{656565} 0.769} & {\color[HTML]{000000} 0.726} & 0.673 & \multicolumn{1}{c|}{0.634} & {\color[HTML]{656565} 0.402} & 0.388 & 0.367 & 0.350 \\
\multicolumn{1}{c|}{} & \multicolumn{1}{l|}{DiscoNet~\cite{Li_2021_NeurIPS}} & {\color[HTML]{656565} 0.791} & 0.766 & 0.746 & \multicolumn{1}{c|}{0.733} & {\color[HTML]{656565} 0.680} & 0.642 & 0.616 & \multicolumn{1}{c|}{0.589} & {\color[HTML]{656565} 0.583} & 0.576 & 0.569 & 0.566 \\
\multicolumn{1}{c|}{\multirow{-4}{*}{\begin{tabular}[c]{@{}c@{}}w/o\\ robust\\ design\end{tabular}}} & \multicolumn{1}{l|}{OPV2V$_{\text{pointpillar}}$~\cite{xu2022opv2v}} & {\color[HTML]{656565} 0.827} & 0.804 & 0.780 & \multicolumn{1}{c|}{0.765} & {\color[HTML]{656565} 0.672} & 0.651 & 0.632 & \multicolumn{1}{c|}{0.625} & {\color[HTML]{656565} 0.553} & 0.547 & 0.540 & 0.538 \\ \hline
\multicolumn{1}{c|}{} & \multicolumn{1}{l|}{MASH~\cite{glaser2021overcoming}} & {\color[HTML]{656565} 0.198} & 0.198 & 0.198 & \multicolumn{1}{c|}{0.198} & {\color[HTML]{656565} 0.384} & 0.384 & {\color[HTML]{000000} 0.384} & \multicolumn{1}{c|}{{\color[HTML]{000000} 0.384}} & {\color[HTML]{656565} 0.244} & 0.244 & 0.244 & 0.244 \\
\multicolumn{1}{c|}{} & \multicolumn{1}{l|}{FPV-RCNN~\cite{9682601}} & {\color[HTML]{656565} 0.840} & 0.568 & 0.278 & \multicolumn{1}{c|}{0.200} & {\color[HTML]{656565} \textbf{0.838}} & 0.617 & 0.352 & \multicolumn{1}{c|}{0.282} & {\color[HTML]{656565} 0.505} & 0.459 & 0.420 & 0.410 \\
\multicolumn{1}{c|}{} & \multicolumn{1}{l|}{V2VNet$_{\text{robust}}$~\cite{pmlr-v155-vadivelu21a}} & {\color[HTML]{656565} 0.854} & 0.848 & 0.837 & \multicolumn{1}{c|}{0.826} & {\color[HTML]{656565} 0.754} & \textbf{0.743} & 0.711 & \multicolumn{1}{c|}{0.676} & {\color[HTML]{656565} 0.486} & 0.483 & 0.478 & 0.475 \\
\multicolumn{1}{c|}{} & \multicolumn{1}{l|}{V2X-ViT~\cite{xu2022v2xvit}} & {\color[HTML]{656565} 0.856} & 0.851 & 0.841 & \multicolumn{1}{c|}{0.823} & {\color[HTML]{656565} 0.726} & 0.708 & 0.673 & \multicolumn{1}{c|}{0.645} & {\color[HTML]{656565} 0.531} & 0.529 & 0.525 & 0.522 \\ \cline{2-14} 
\multicolumn{1}{c|}{\multirow{-5}{*}{\begin{tabular}[c]{@{}c@{}}w/\\ robust\\ design\end{tabular}}} & \multicolumn{1}{l|}{Ours} & {\color[HTML]{656565} \textbf{0.912}} & {\color[HTML]{000000} \textbf{0.900}} & {\color[HTML]{000000} \textbf{0.889}} & \multicolumn{1}{c|}{{\color[HTML]{000000} \textbf{0.868}}} & {\color[HTML]{656565} 0.765} & 0.742 & {\color[HTML]{000000} \textbf{0.711}} & \multicolumn{1}{c|}{{\color[HTML]{000000} \textbf{0.684}}} & {\color[HTML]{656565} \textbf{0.604}} & {\color[HTML]{000000} \textbf{0.588}} & {\color[HTML]{000000} \textbf{0.579}} & {\color[HTML]{000000} \textbf{0.570}} \\ \hline
\end{tabular}%
}
\caption{Detection performance on OPV2V~\cite{xu2022opv2v}, V2X-Sim 2.0~\cite{Li_2021_RAL} and DAIR-V2X~\cite{yu2022dairv2x} datasets with pose noises following Gaussian distribution in the
testing phase. All models are trained on pose noises following with $\sigma_t=0.2m$, $\sigma_r=0.2$°. Experiments show that $\mathtt{CoAlign}$ holds the best resistance to localization error under various noise levels. }
\label{tab:performance_all}
\vspace{-5mm}
\end{table*}

\subsection{Multiscale Feature Fusion}
\label{sec:fusion}
After the spatial alignment, each agent aggregates other agents' collaboration information and obtain a more informative feature. However, even after the relative pose correction, the misalignment between feature maps might still exist. To further mitigate the effect of pose noises, we adopt a multiscale fusion method, which fuses features at multiple spatial scales. Features at the finer scale can provide more detailed geometric and semantic information while the features at the coarser scale are less sensitive to the pose error. The multiscale structure can take both advantages and produce both informative and robust features. 

Mathematically,  let $\mathbf{F}_{j \rightarrow i}^{(\ell)}$ be the collaboration feature at the $\ell$th spatial scale. Then, the multiscale fusion works as 
\begin{subequations}
\begin{eqnarray*}
    \label{eq:fusion_initialization}
    \mathbf{F}_{j \rightarrow i}^{(1)} & = & \mathbf{M}_{j \rightarrow i}, j=1,2,\cdots N,
    \\
    \label{eq:fusion_layer}
    \mathbf{F}_{j \rightarrow i}^{(\ell+1)} & = & g_{\ell}(\mathbf{F}^{(\ell)}_{j \rightarrow i}), \ell=1,2, \cdots,L, 
    \\
    \label{eq:fusion_att}
    \mathbf{F}^{(\ell)}_i & = & \text{Fuse}(\{ \mathbf{F}^{(\ell)}_{j \rightarrow i} \}_{j=1,2, \cdots,N}), 
    \\
    \label{eq:fusion_concat}
    \mathbf{F}'_i & = & \text{Cat}([\mathbf{F}^{(1)}_i, u_2(\mathbf{F}^{(2)}_i),\cdots, u_L(\mathbf{F}^{(L)}_i)]), 
\end{eqnarray*}
\end{subequations}
where $g_{\ell}(\cdot)$ is the $\ell$th residual layer with downsampling by $2$, Fuse$(\cdot)$ is a attention operation along the agent dimension, $u_{\ell}(\cdot)$ is an upsampling operator for the $\ell$th scale and Cat$(\cdot)$ is a concatenation operator along the feature dimension. Fused feature $\mathbf{F}'_i$ will be decoded into final detection.



\begin{table}[]
\centering
\resizebox{0.48\textwidth}{!}{%
\begin{tabular}{c|l|cccccccc}
\hline
\multicolumn{1}{l|}{} & \multicolumn{1}{c|}{Method/Metric} & \multicolumn{4}{c}{\textbf{AP@0.5 $\uparrow$}} & \multicolumn{4}{c}{\textbf{AP@0.7 $\uparrow$}} \\ \hline
\multicolumn{1}{l|}{} & \multicolumn{1}{c|}{\begin{tabular}[c]{@{}c@{}}Noise Level\\ $b_t/b_r$ (m/°)\end{tabular}} & {\color[HTML]{656565} 0.0/0.0} & 0.2/0.2 & 0.4/0.4 & \multicolumn{1}{c|}{0.6/0.6} & {\color[HTML]{656565} 0.0/0.0} & 0.2/0.2 & 0.4/0.4 & 0.6/0.6 \\ \hline
 & F-Cooper & {\color[HTML]{656565} 0.734} & 0.716 & 0.694 & \multicolumn{1}{c|}{0.681} & {\color[HTML]{656565} 0.559} & 0.552 & 0.541 & 0.530 \\
 & V2VNet & {\color[HTML]{656565} 0.665} & 0.639 & 0.606 & \multicolumn{1}{c|}{0.585} & {\color[HTML]{656565} 0.401} & 0.379 & 0.356 & 0.342 \\
 & DiscoNet & {\color[HTML]{656565} 0.736} & 0.718 & 0.700 & \multicolumn{1}{c|}{0.688} & {\color[HTML]{656565} 0.583} & 0.574 & 0.567 & 0.563 \\
\multirow{-4}{*}{\begin{tabular}[c]{@{}c@{}}w/o\\ robust\\ design\end{tabular}} & Self-Att & {\color[HTML]{656565} 0.733} & 0.718 & 0.701 & \multicolumn{1}{c|}{0.692} & {\color[HTML]{656565} 0.553} & 0.546 & 0.538 & 0.536 \\ \hline
 & MASH & {\color[HTML]{656565} 0.400} & 0.400 & 0.400 & \multicolumn{1}{c|}{0.400} & {\color[HTML]{656565} 0.244} & 0.244 & 0.244 & 0.244 \\
 & FPV-RCNN & {\color[HTML]{656565} 0.654} & {\color[HTML]{000000} 0.609} & 0.564 & \multicolumn{1}{c|}{0.553} & {\color[HTML]{656565} 0.504} & 0.436 & 0.417 & 0.431 \\
 & V2VNet(robust) & {\color[HTML]{656565} 0.660} & 0.653 & 0.640 & \multicolumn{1}{c|}{0.632} & {\color[HTML]{656565} 0.486} & 0.481 & 0.476 & 0.469 \\
\multirow{-4}{*}{\begin{tabular}[c]{@{}c@{}}w/\\ robust\\ design\end{tabular}} & V2X-ViT & {\color[HTML]{656565} 0.705} & 0.695 & 0.680 & \multicolumn{1}{c|}{0.667} & {\color[HTML]{656565} 0.531} & 0.527 & 0.521 & 0.517 \\ \hline
\multicolumn{1}{l|}{} & Ours & {\color[HTML]{656565} \textbf{0.746}} & \textbf{0.733} & {\color[HTML]{000000} \textbf{0.707}} & \multicolumn{1}{c|}{{\color[HTML]{000000} \textbf{0.689}}} & {\color[HTML]{656565} \textbf{0.604}} & {\color[HTML]{000000} \textbf{0.585}} & {\color[HTML]{000000} \textbf{0.573}} & {\color[HTML]{000000} \textbf{0.566}} \\ \hline
\end{tabular}%
}
\caption{Detection performance on DAIR-V2X with Laplace pose noises in the testing phase. Models are trained on Gaussian pose noises with $\sigma_t=0.2m, \sigma_r=0.2^\circ$. $\mathtt{CoAlign}$ generalizes well to handle unexpected noises. }
\vspace{-5mm}
\label{dairv2x-result-laplace}
\end{table}

\section{Experimental Results}
 We validate $\mathtt{CoAlign}$ in the task of collaborative LiDAR-based 3D object detection.
 \vspace{-3mm}
\subsection{Dataset}
\textbf{V2X-Sim 2.0~\cite{Li_2021_RAL}} is a vehicle-to-everything collaborative perception dataset, co-simulated by SUMO~\cite{krajzewicz2012recent} and Carla~\cite{dosovitskiy2017carla}, including 10K frames of 3D LiDAR point clouds and 501K 3D boxes. 
We follow ~\cite{Li_2021_RAL} and set the detection range as $x\in [-32m,32m],  y \in [-32m, 32m]$. \textbf{OPV2V~\cite{xu2022opv2v}} is a vehicle-to-vehicle collaborative perception dataset, co-simulated by OpenCDA~\cite{xu2021opencda} and Carla~\cite{dosovitskiy2017carla}. It includes 12K frames of 3D LiDAR point clouds and RGB images with 230K annotated 3D boxes. We follow ~\cite{xu2022opv2v} and set the detection range as $x\in [-140m, 140m], y\in[-40m, 40m]$. The original paper considers a 2-round communication setting where each agent shares pose first and encodes feature map in receiver's coordinate by transforming its point cloud to mitigate the discretization issue. It leads to higher performance but is less practical as each agent's computational cost is expensive when the number of collaborators goes up. Instead, here we make each agent transmit the same feature map to all collaborators in 1-round communication. \textbf{DAIR-V2X~\cite{yu2022dairv2x}} is a real-world collaborative perception dataset. It contains 9K cooperative frames with one vehicle and one road-side unit. We expand the detection range to $x\in [-100 m, 100m], y\in[-40m, 40m]$. Originally DAIR-V2X does not label objects outside the camera's view, we complemented the missing label to cover 360-degree detection.

\vspace{-1mm}
\subsection{Implementation Details}
To simulate pose errors, we add Gaussian noise $\mathcal{N}(0,\sigma_t)$ on $x, y$ and $\mathcal{N}(0, \sigma_r)$ on $\theta$ during the full training phase, where $x,y,\theta$ are 2D centers and yaw angle of accurate global poses.  To train the detection model, we set $\alpha_{reg} = 2,  \alpha_{center} = 0.25, \alpha_{\theta}=0.125, \lambda_{elu}=0.01, c=1$. We use PointPillars~\cite{lang2019pointpillars} with the grid size $(0.4m, 0.4m)$ as the encoder. To optimize over the agent-object pose graph, we adopt the general graph optimization(g2o)~\cite{grisetti2011g2o} library. Both edges and vertices are in $SE(2)$. Here we select the dense solver for 
incremental equations and Levenberg-Marquartdt algorithms\cite{levenberg1944method} for iterative optimiztion with max iteration time 1000. For multiscale feature fusion, the residual layer number is 2 and channel numbers are $(128, 256)$ respectively.  We use Adam~\cite{kingma2014adam} with learning rate 0.001 for single detection training and 0.002 for intermediate fusion training. We set batchsize 4 and epoch number 30. $\mathtt{CoAlign}$ converges within 10 hours on OPV2V dataset with one RTX 3090, while other methods like V2X-ViT take more than one day.

\subsection{Quantitative Results}
To validate the overall 3D detection performances in presence of pose errors, we compare the proposed $\mathtt{CoAlign}$ with a series of previous methods with and without pose-robust designs. Table~\ref{tab:performance_all} shows the average precisions (AP) at Intersection-over-Union (IoU) threshold of 0.5 and 0.7 in OPV2V, V2X-Sim 2.0 and DAIR-V2X datasets. We see that $\mathtt{CoAlign}$ significantly outperforms the previous methods in various noise levels across all three datasets and the leading gap is bigger when the noise level is higher.

\begin{figure}[t]
\centering
\includegraphics[width=\linewidth]{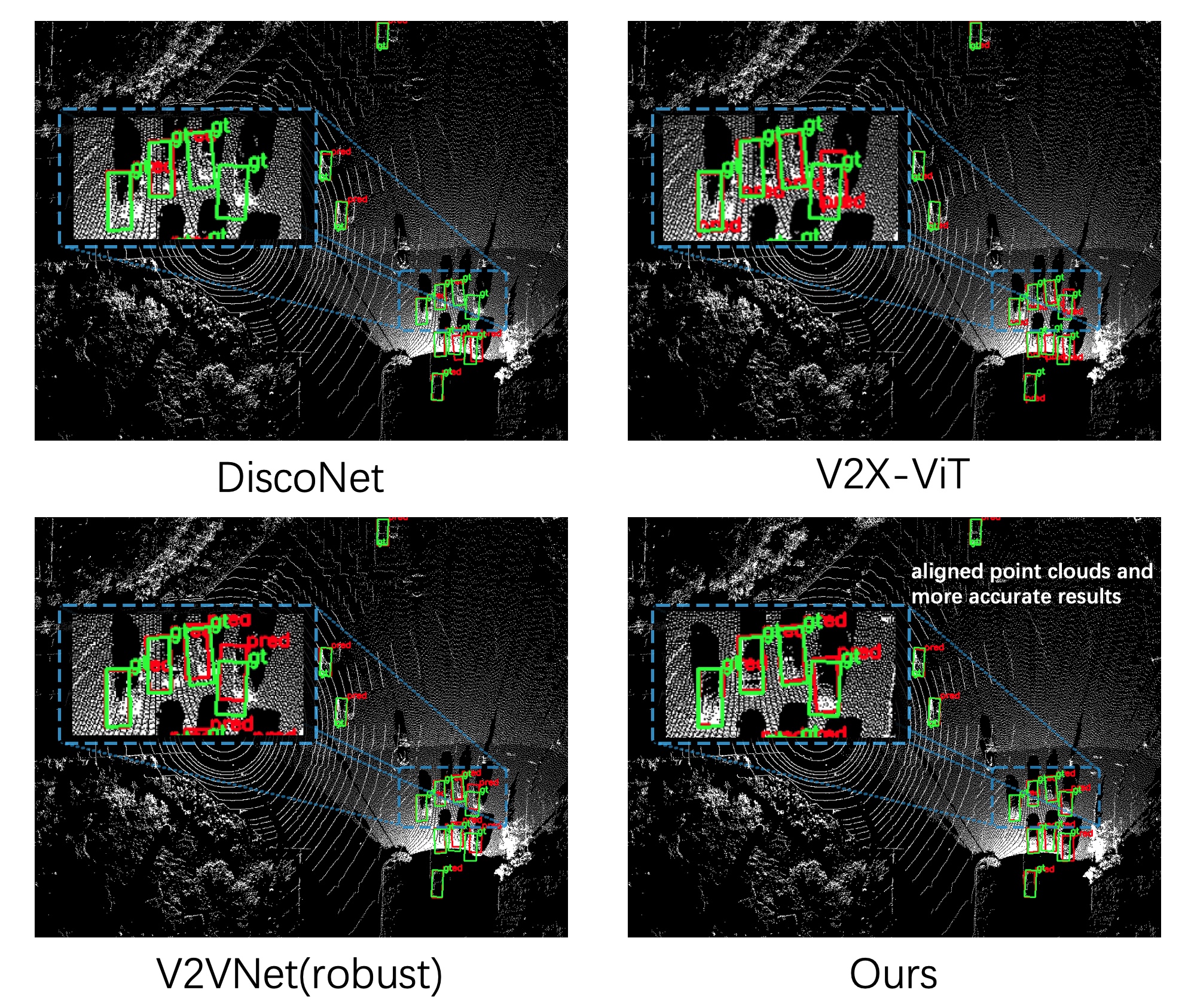}
\caption{Visualization of detected boxes in DAIR-V2X dataset. Green boxes are \textcolor{green}{ground-truth} while red ones are \textcolor{red}{detection}. $\mathtt{CoAlign}$ achieves much more precise detection.}
\vspace{-5mm}
\label{fig:detection}
\end{figure}

\begin{figure}[h]
\centering
\subfloat[Boxes w/o correction ]{\includegraphics[width=0.48\linewidth]{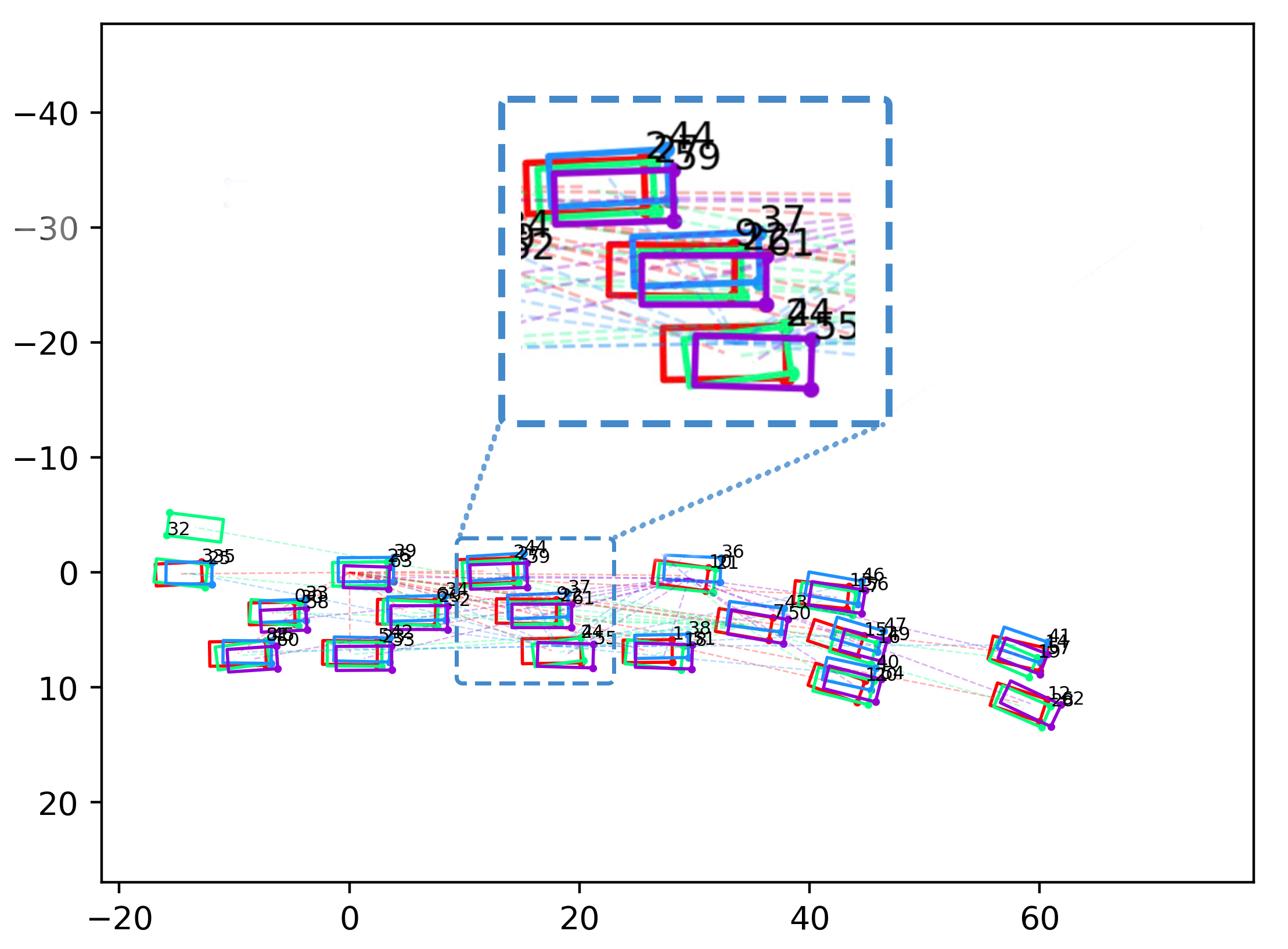}}
\subfloat[Boxes w/ correction ]{\includegraphics[width=0.48\linewidth]{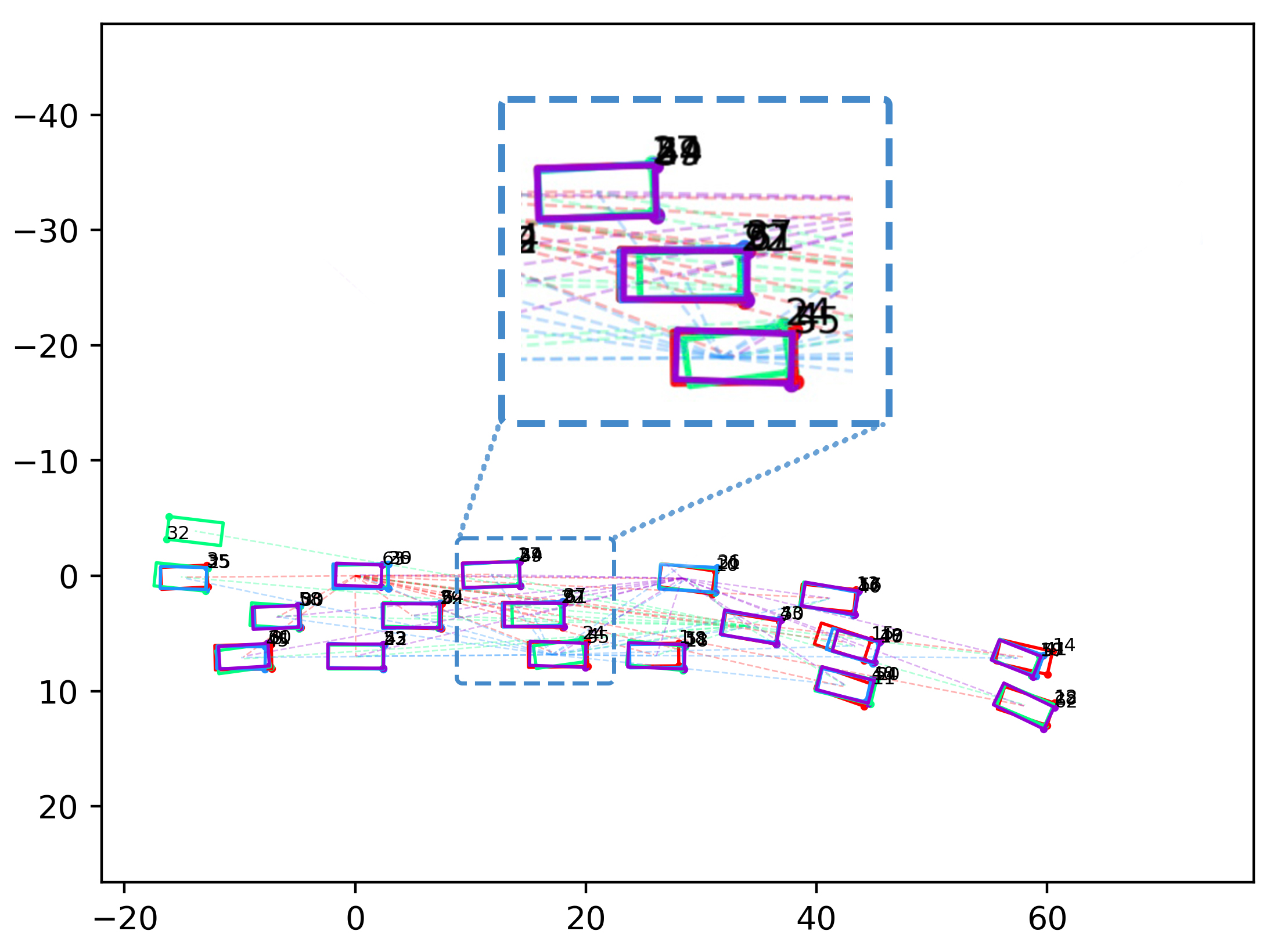}}
\caption{Agent-object pose graph optimization can well align boxes detected by multiple agents in OPV2V dataset.}
\label{fig:pg-before-after}
\vspace{-5mm}
\end{figure}

\subsection{Qualitative Results}
Figure~\ref{fig:detection} visualizes the detected boxes on DAIR-V2X dataset. We see that the boxes detected by $\mathtt{CoAlign}$ are much more precise than those of other methods. Figure~\ref{fig:pg-before-after} visualize the global box position before and after agent-object pose graph optimization on OPV2V dataset. Different colors represent predictions from different agent views. We see that box positions achieve great consistency after alignment.
\vspace{-2mm}

\begin{table}[]
\centering
\resizebox{0.48\textwidth}{!}{%
\begin{tabular}{c|ccc|cccc}
\hline
\multicolumn{1}{l|}{} & \multicolumn{3}{c|}{Modules} & \multicolumn{4}{c}{\textbf{AP@0.7 $\uparrow$}} \\ \cline{2-4}
\begin{tabular}[c]{@{}c@{}}Collab-\\ oration\end{tabular} & \begin{tabular}[c]{@{}c@{}}Agent-Object \\ Pose Graph\end{tabular} & Uncertainty & \begin{tabular}[c]{@{}c@{}}Intermediate \\ Fusion\end{tabular} & 0.0/0.0 & \multicolumn{1}{l}{0.2/0.2} & \multicolumn{1}{l}{0.4/0.4} & \multicolumn{1}{l}{0.6/0.6} \\ \hline
\multicolumn{1}{l|}{} &  &  &  & {\color[HTML]{656565} 0.703} & 0.703 & 0.703 & 0.703 \\
 &  & \checkmark &  & {\color[HTML]{656565} 0.730} & 0.730 & 0.730 & 0.730 \\ \hline
\checkmark &  &  & / & {\color[HTML]{656565} 0.907} & 0.490 & 0.275 & 0.239 \\
\checkmark & \checkmark &  & / & {\color[HTML]{656565} 0.899} & 0.814 & 0.751 & 0.657 \\
\checkmark & \checkmark & \checkmark & / & {\color[HTML]{656565} 0.903} & 0.818 & 0.758 & 0.672 \\
\checkmark &  &  & Single-scale & {\color[HTML]{656565} 0.824} & 0.789 & 0.766 & 0.757 \\
\checkmark &  &  & Multi-scale & {\color[HTML]{656565} \textbf{0.914}} & 0.860 & 0.799 & 0.768 \\
\checkmark & \checkmark & \textit{} & Multi-scale & {\color[HTML]{656565} 0.910} & 0.897 & 0.886 & 0.865 \\
\checkmark & \checkmark & \checkmark & Multi-scale & {\color[HTML]{656565} 0.912} & {\color[HTML]{000000} \textbf{0.900}} & {\color[HTML]{000000} \textbf{0.889}} & {\color[HTML]{000000} \textbf{0.868}} \\ \hline
\end{tabular}%
}
\caption{Ablation studies on OPV2V dataset. All technique modules benefit 3D collaborative object detection. In Intermediate Fusion column, $/$ is late fusion.}
\vspace{-3mm}
\label{tab:ablation-studies}
\end{table}
\begin{figure}[!t]
\centering
\includegraphics[width=0.99\linewidth]{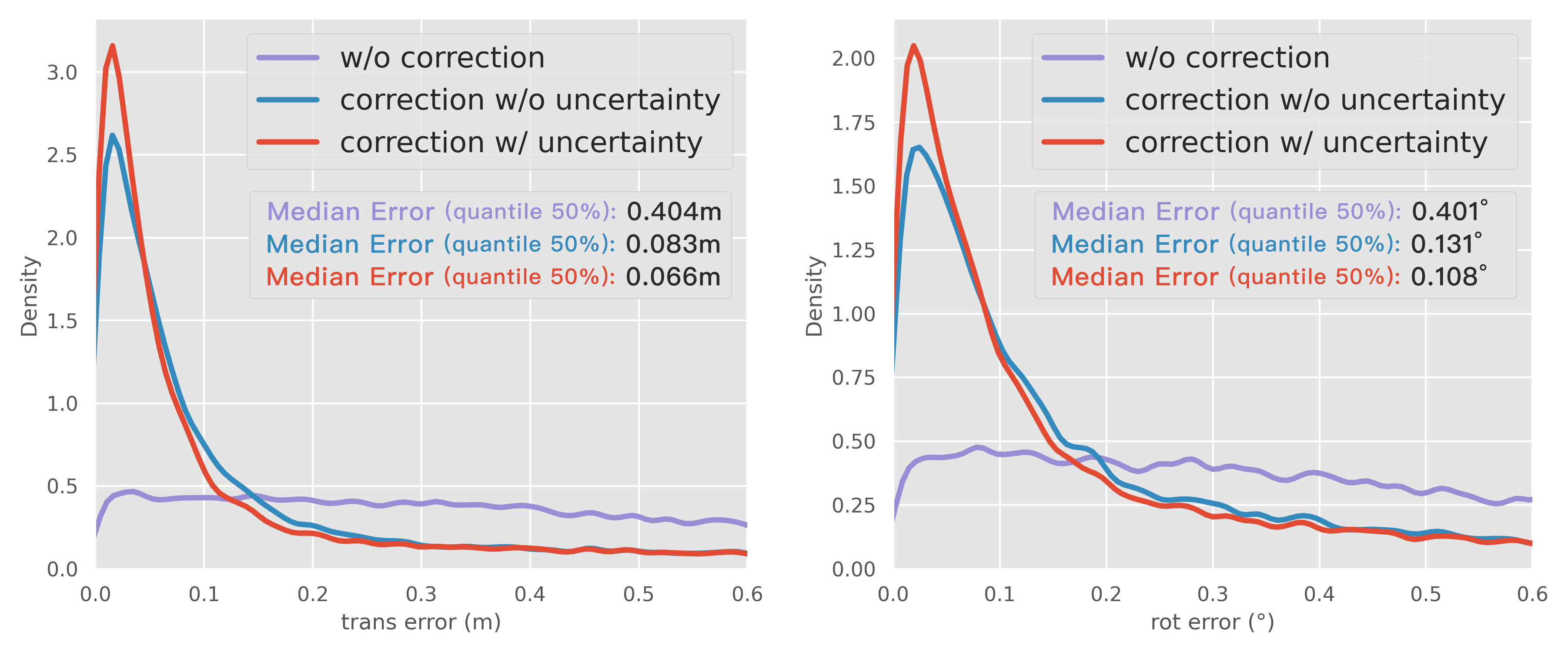}
\caption{Probability density function (PDF) of relative pose error distribution on OPV2V dataset when noise level is $\sigma_t/\sigma_r$=0.6m/0.6°. We see that agent-object pose graph optimization significantly reduces the relative pose error and uncertainty modeling can further refine the alignment.}
\vspace{-5mm}
\label{fig:localization-error}
\end{figure}

\subsection{Ablation Studies}
To validate the efficacy of each module, Table~\ref{tab:ablation-studies} shows the AP at IoU 0.7 at various noise levels on OPV2V dataset. Note that i) agent-object pose graph promotes the stability of collaboration,  ii) multi-scale fusion outperforms single-scale fusion with powerful hierarchical structure, iii) uncertainty modeling refines the correction, under all noise levels.

To validate the agent-object pose alignment module, Figure~\ref{fig:localization-error} plots the distributions of relative pose errors across all samples. The performance is better when the distribution is close to the delta distribution centered at zero. Agent-object pose alignment can significantly decrease the pose error and uncertainty modeling can further enhance the error correction capability. The median errors are reduced to 25\% of the original.


\section{Conclusion}
This paper proposes a new hybrid collaboration framework $\mathtt{CoAlign}$ for robust 3D object detection. The proposed agent-object pose graph optimization empowers $\mathtt{CoAlign}$ to handle arbitrary pose errors without any accurate pose supervision. $\mathtt{CoAlign}$ can not only reduce the relative pose noises significantly, but improve the detection ability and robustness. Furthermore, $\mathtt{CoAlign}$ does not rely on certain data modality and can be applied to camera-based 3D detection as well. In future works, we will extend our method on multimodal data.


\section{Acknowledge}
This work was supported in part by the National Natural Science Foundation of China under Grant 62171276, in part by the Science and Technology Commission of Shanghai Municipal under Grant 21511100900.

\bibliographystyle{IEEEtran}
\bibliography{IEEEfull}

\end{document}